\documentclass[runningheads]{llncs}
\usepackage{graphicx}
\usepackage{amsmath}
\usepackage{amsfonts}
\usepackage{multirow}
\usepackage{pifont}
\begin{document}
\title{Rapid treatment planning for low-dose-rate prostate brachytherapy with TP-GAN \thanks{This work was supported by the Canadian Institutes of Health Research (CIHR).}}
\titlerunning{Rapid treatment planning for LDR-PB with TP-GAN}
\author{Tajwar Abrar Aleef\inst{1} \and
Ingrid T. Spadinger\inst{2} \and
Michael D. Peacock\inst{2} \and
Septimiu E. Salcudean\inst{3} \and
S. Sara Mahdavi \inst{2}}
\authorrunning{Aleef, T.A. et al.}
\institute{School of Biomedical Engineering, University of British Columbia, Vancouver, Canada \and
BC Cancer - Vancouver Centre, Vancouver, Canada \and
Department of Electrical and Computer Engineering, University of British Columbia, Vancouver, Canada\\ 
\email{tajwaraleef@ece.ubc.ca}}
\maketitle              % typeset the header of the contribution
\begin{abstract}
Treatment planning in low-dose-rate prostate brachytherapy (LDR-PB) aims to produce arrangement of implantable radioactive seeds that deliver a minimum prescribed dose to the prostate whilst minimizing toxicity to healthy tissues. There can be multiple seed arrangements that satisfy this dosimetric criterion, not all deemed `acceptable' for implant from a physician's perspective. This leads to plans that are subjective to the physician's/centre's preference, planning style, and expertise. We propose a method that aims to reduce this variability by training a model to learn from a large pool of successful retrospective LDR-PB data ($961$ patients) and create consistent plans that mimic the high-quality manual plans. Our model is based on conditional generative adversarial networks that use a novel loss function for penalizing the model on spatial constraints of the seeds. An optional optimizer based on a simulated annealing (SA) algorithm can be used to further fine-tune the plans if necessary (determined by the treating physician). Performance analysis was conducted on $150$ test cases demonstrating comparable results to that of the manual prehistorical plans. On average, the clinical target volume covering $100\%$ of the prescribed dose was $98.9\%$ for our method compared to $99.4\%$ for manual plans. Moreover, using our model, the planning time was significantly reduced to an average of $2.5$ mins/plan with SA, and less than $3$ seconds without SA. Compared to this, manual planning at our centre takes around $20$ mins/plan.

\end{abstract}

\section{Introduction}
Low-dose-rate prostate brachytherapy (LDR-PB) is considered an effective curative treatment for men with localized prostate cancer (PCa) \cite{stish2018low}. In LDR-PB, a standard needle template is used to guide and place permanent radioactive seeds transperineally into the prostate through needles \cite{stish2018low,yu1999permanent}. Before the actual implant, expert planners manually determine the optimal distribution of the seeds that deliver a prescribed dose to the target anatomy while minimizing toxicity to other surrounding tissues. Several seed arrangements can fulfil dosimetric constraints and clinical guidelines, although not all these plans are deemed acceptable for implant by the physician, due to factors such as pelvic arch interference in the patient and the physician's personal preference on what might make a plan easier to deliver. The choice of selecting optimal plans can hence be subjective where the outcome depends highly on the expertise and preference of the planner. Furthermore, this is a time-consuming task for brachytherapy clinicians. For automating this planing procedure, which can also benefit real-time intra-operative LDR-PB planning, many approaches have been proposed. Common methods focus on meeting the dose-volume criteria with limited constraints on the needle or seed locations (which can determine if a plan is implantable or not). These include various optimization techniques such as: mixed-integer linear programming \cite{MILP}, inverse treatment planning with compressed sensing \cite{compressedsensing}, genetic algorithms \cite{genetic}, and fast simulated annealing \cite{sa}. Most of these techniques are very sensitive to their initialization, and the associated optimization costs have multiple minima and a huge search space. The use of machine learning approaches aiming to produce similar-looking plans learned from respective centre's data is limited in the literature \cite{saman,nicolae}. Both these methods initialize a prior plan based on the training data followed by the main optimization step. To initiate the priors from their respective training database, \cite{saman} uses a joint Sparse Dictionary Learning approach to learn the relationship between target volumes and seed plans while \cite{nicolae} uses a feature extraction \& matching technique to look for similar plans. Both methods use a limited set of features that are selected manually and they rely heavily on the optimization step. Hence, plan characteristics will mostly be those defined within the objective function with less chance of learning plan features that cannot be mathematically expressed. Indeed, \cite{saman} reports a significant drop in performance when its optimizer is removed.\\

In this work, we propose a deep learning method that learns from a large set of $961$ retrospective successful clinical plans to produce treatment plans for LDR-PB in an end-to-end manner. Our model is based on supervised learning where it trains a large set of learnable parameters, capturing implicit clinical features from the data automatically. The proposed model predicts seed plans directly using volumetric data from the anatomy (easily available) and constraints from the needle space (as per standard guidelines). To achieve this, we use a novel loss function that incorporates additional spatial constraints for the seed placements. As supervised learning technique is limited to what it has observed in the training dataset, and the problem can have multiple solutions, we also provide an option to fine-tune the results further when required using a simulated annealing (SA) based optimizer. Using several key plan quality metrics, we compare our method with existing and common approaches of automatic plan generation. A further ablation study is conducted to validate the need for the different components of our model. To our best knowledge, this is the first technique using deep learning methods to learn this multi-objective mapping for LDR-PB in an end-to-end approach.

\section{Methods}
\subsection{Dataset}
This study used retrospective clinical data of $961$ patients taken from BC Cancer (Vancouver, BC, Canada) with institutional ethics approval. This data includes annotations of the clinical target volumes (CTV, i.e. the prostate) and the planning target volumes (PTV, i.e. the CTV plus a predefined margin). For every patient, $2\text{-}4$ plan variations were created by the medical physicists from which one was selected for implantation by the treating physician. These plans consist of seeds located on a $5mm$ spaced grid of size $11 \times 13\times \#axial~image~planes$, where $11 \times 13$ is the needle template size (see Fig.~\ref{fig:2}). The data includes prostate volumes ranging from $20\text{-}70$ cc, all receiving standard I-$125$ monotherapy with a prescribed dose of $144$ Gy. We used $711$ cases for training, $100$ cases for validation, and the remaining $150$ cases for testing. 

\subsection{Seed planning with TP-GAN}
We propose a cGANs based model that we refer to as ``TP-GAN" (treatment planning with GAN). Multiple seed arrangements can deliver the prescribed dose to the volume of the gland. However, plans are decided not only based on dose coverage but also considering a pool of guidelines that are considered best practice for LDR-PB. Moreover, the expertise of the planners and treating physicians also determines which plan will eventually be selected for the implant. With such a degree of variation, there can be multiple solutions satisfying the planning guidelines, and hence our clinical dataset does not embody a straightforward solution path that can be depicted only from the target volume data. To alleviate this, we narrow down the search space by putting a constrain on the needle space. Our group previously showed that centre specific needle plans can be automatically \& reliably generated given the target volume data (PTV, CTV) \cite{ipcai}. The problem hence becomes an image-to-image mapping task where the inputs are the volumetric data (PTV, CTV) with the automatically generated needle plan and the output is the corresponding seed plan.\\ 

The volumetric data is firstly represented as binary masks where all pixels within the margins are set to `1' and the rest of the background are set to `0'. All the volumetric data and plans are aligned to a reference needle template. The number of axial image planes is set to 14 to cover the maximum prostate length of $14 \times 5mm$, with axial planes zero-padded for shorter prostates. Likewise, the needle and seed plans were also converted to a binary matrix following the needle template grid where a pixel with `1' means a presence of a seed/needle in that grid coordinate. The top row of the template is removed from the output as our centre does not accept seed placements there. This makes the dimension of the seed matrix $10 \times 13 \times 14$ and the needle matrix $10\times13$. A weighted distance transform was then performed on the binary volumetric and plan data to provide the network with further distance-based information \cite{karimi2019reducing}. As the needle plans are 2D and are part of the input, they were resized and stacked to match with the dimension of the 3D volumetric data. All PTV, CTV and needle plans were axially resized and then stacked in channels to form the input with dimension $64 \times 64 \times 14 \times 3$. Data augmentation increases the robustness of deep networks and reduces the chance of over-fitting. However, typical geometric or image-based augmentation technique can't be used for our problem as any change in the input space doesn't imply the same change to the output space. Since the plans in our dataset are symmetric, we utilize the following augmentation technique to double the data for training: the data is vertically split around the center of the template and the left and right sides are treated as two independent samples. After augmentation, the dimension of the input and output data becomes $64 \times 32 \times 14$ and $10 \times 6 \times 14$. \\ 

\begin{figure}[t]
%left bottom right up: clip
\includegraphics[width=1.0\textwidth, trim={4cm 3.5cm 4.8cm 2.5cm},clip]{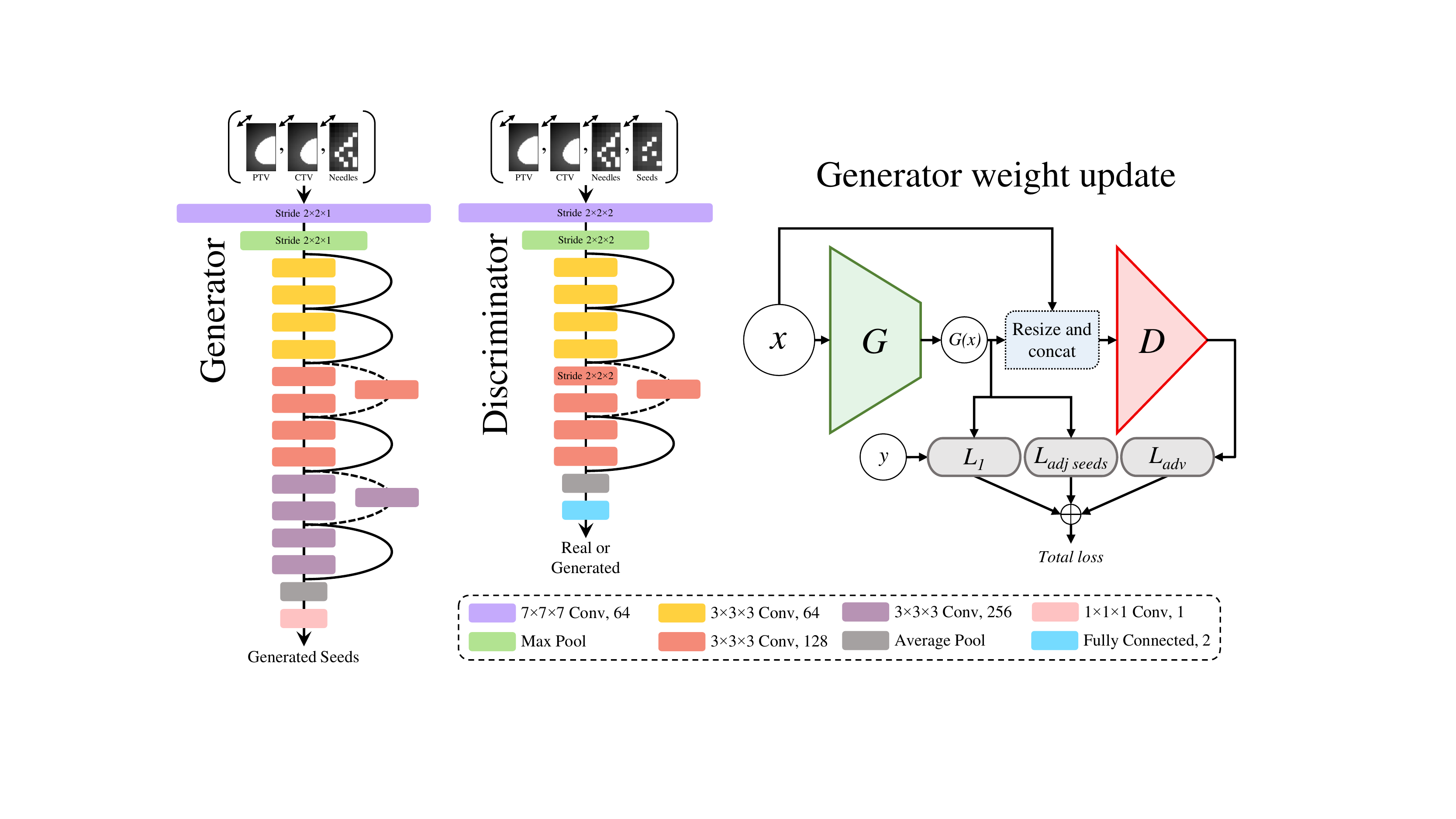}
\caption{Architectures of the generator ($G$) and discriminator ($D$) of TP-GAN. Figure legend provides kernel and filter sizes of the different layers. The diagram on the right shows how the different losses are calculated for updating the weight of $G$. The input of the model (PTV, CTV, and needle plan) is given by $x$ and the output of the model is a seed plan given by $G(x)$. $y$ is the corresponding manual plan. The three light grey block shows the three losses used to optimize $G$.}
\label{fig:architechture}       % Give a unique label
\end{figure}

We design the TP-GAN model based on \cite{pix2pix}- which showed results with great generalizability in terms of learning paired image-to-image transformations for a plethora of applications. TP-GAN consists of a generator ($G$) that learns to encode the input space ($x$) to a seed plan ($G(x)$) and a discriminator ($D$) that learns to recognize between real ($y$) and generated plans ($G(x)$). The weights of the two networks are updated alternatively \cite{Gans}. The architecture of $G$ is designed out of a ResNet architecture \cite{resnet} with 14 layer depth that encodes 4D inputs to 3D seed plans. $D$ is also based on a ResNet architecture with a depth of 10 layer which takes the pair of inputs \& outputs and predicts if it is generated or real. A resize and concatenation block is used to resize the real/generated output and stack it with $x$ in channels for $D$. Details of the architectures and the loss calculation for updating weights of $G$ are given in Fig.~\ref{fig:architechture}. \\

To minimize hot spots (i.e. high dose regions) in the prostate, a general preference is to avoid implanting adjacent seeds. Although introducing needle plans to the input constrains the model to produce seeds within those needles, without further regularization, the model tends to produce plans with neighbouring adjacent seeds. To mitigate this problem, in addition to the $Adversarial$ and $L_1$ loss from \cite{pix2pix}, we introduce a new loss function that imposes a penalty for predicting adjacent seeds during training. \\

\medskip\noindent The $Adversarial$, $L_1$, and $Adjacent~seed$ losses are given by:

\begin{equation}
\mathcal{L}_{adv}(G,D) = \mathbb{E}_{x,y}[\log D(x,y)] + \mathbb{E}_{x}[\log(1-D(x,G(x)))] \label{eq:1}
\end{equation}

\begin{equation}
\mathcal{L}_{L_1}(G) = \mathbb{E}_{x,y}[||y - G(x)||_{1}] \label{eq:2}
\end{equation}

\begin{equation}
\mathcal{L}_{adj\:seeds}(G) = \mathbb{E}_{x}[\sum(\max(0,(G(x) * k) - 5)] \label{eq:3}
\end{equation}

\begin{equation*}
where,~k = 
\begin{bmatrix}
\begin{bmatrix}
0 & 0 & 0\\ 
0 & 1 & 0\\
0 & 0 & 0
\end{bmatrix}
\begin{bmatrix}
0 & 1 & 0\\ 
1 & 7 & 1\\
0 & 1 & 0
\end{bmatrix}
\begin{bmatrix}
0 & 0 & 0\\ 
0 & 1 & 0\\
0 & 0 & 0
\end{bmatrix}

\end{bmatrix}
\end{equation*}

\medskip\noindent In the $Adjacent~seed$ loss (Equation \ref{eq:3}), \textit{k} is a $3\times3\times3$ kernel used to find the presence of adjacent seeds in the prediction. The adjacent seeds can be localized by convolving \textit{k} with the output. If the prediction of the model is binary, this convolution will set any adjacent coordinates to greater than 7. However, since the output is a probability in the range between $[0,1]$, we lower this threshold and set all values lower than 5 to negative. This 5 is a hyperparameter that was selected during the tuning phase. To only keep positive pixels where adjacent seeds are observed, a $ReLu$ function ($max(0,X)$) is used that sets the negative values to zero. Now all remaining pixels correspond to the adjacent seeds which are then summed to form this loss.

\medskip\noindent The full objective function of TP-GAN then becomes: 
\begin{equation}
\begin{aligned}
\mathcal{L}(G,D) =  \alpha(\arg\min_G\max_D{\mathcal{L}_{adv}(G,D))
+ \beta\mathcal{L}_{L1}(G)
+ \alpha\mathcal{L}_{adj\:seeds}(G)}  \label{eq:4}
\end{aligned}
\end{equation}

\medskip\noindent where, the weights of the different losses were tuned to $\alpha$ =  \( \frac{1}{3} \) and $\beta$ =\( \frac{2}{3} \).\\

Keras and Tensorflow were primarily used to train, validate, and test our model. For both the $G$ and $D$, Adam optimizer with a learning rate of $10^{-5}$ and momentum of $\beta_1 = 0.5$ and $\beta_2 = 0.99$ were used. The model was trained for 1000 epochs with a batch size of 16 samples on a single NVIDIA Tesla V100 GPU (16GB) which took about 25 hrs to train. Weights of the model with the lowest validation loss were used for the evaluation of the results. Code of our implementation is available at: \textit{https://github.com/tajwarabraraleef/TP-GAN}

\subsection{Post processing stage and fine-tuning with Simulated Annealing}
The generated seed plans are then passed to a post-processing step that checks for any remaining adjacent seeds in the output. If found, it attempts to first relocate the seeds; otherwise, it removes them. A further uniformization stage is then used to improve the uniformity of seed distribution between planes automatically when needed \cite{ipcai}. These stages are computationally inexpensive and don't increase the planning duration. An optional, but beneficial step is the use of an SA optimizer based on \cite{sarasa,sa} initialized using results of the post-processing step. SA fine-tunes the results by further enforcing some dosimetric constraints. Hence planning benefits from a combination of learned factors through the training set and explicit constraints determined by clinical guidelines. As we show, this step particularly reduces the unnecessary urethral dose. Furthermore, initializing the SA with our method provides it with a solution that is already close to the global minimum.

\section{Results}
\subsection{Performance analysis}
The quality of the plans was assessed using standard dosimetric parameters including: the target volume dose coverage (PTV \textit{V}100\%, PTV \textit{V}150\%, CTV \textit{V}100\%, CTV \textit{V}150\%), the urethra \& rectum (OAR: organs at risk) exposure (URE \textit{V}150\% \& REC \textit{V}50\%), and source usages indicated by the number of seeds \& needles used in the plan. Here, ~\textit{V}$x\%$ indicates the percentage of the target volume (PTV, CTV, URE, or REC) receiving $x\%$ of the minimum prescribed dose.  Table \ref{tab:1} lists the mean and standard deviations of the key quality metrics on the 150 test cases for a number of plan generation techniques which includes: 1) SA initialized using Seattle based planning method ($SA_{seattle}$) \cite{seattlestyle}, 2) joint Sparse Dictionary Learning approach from \cite{saman} ($jSDL$), 3) SA initialized with generated needle plans from \cite{ipcai} ($SA_{genN}$), 4) Just TP-GAN ($TP\text{-}GAN$), 5) TP-GAN with SA tuning ($TP\text{-}GAN+SA$), and 6) Manual prehistoric plans ($Actual$). \\

\begin{table}[t]
\centering
\caption{Comparison of the mean and standard deviation of the key plan quality metrics for different techniques on the test set. \#N and \#S indicates the count of needles and seeds used. Results from our method are in bold.}
\label{tab:1}
\resizebox{\textwidth}{!}{%
\begin{tabular}{c@{\hskip 0.05in}|@{\hskip 0.05in}c@{\hskip 0.05in}|@{\hskip 0.05in}c@{\hskip 0.05in}|@{\hskip 0.05in}c@{\hskip 0.05in}|@{\hskip 0.05in}c@{\hskip 0.05in}|@{\hskip 0.05in}c@{\hskip 0.05in}|@{\hskip 0.05in}c@{\hskip 0.05in}|@{\hskip 0.05in}c@{\hskip 0.05in}|@{\hskip 0.05in}c}
\textbf{} & \textbf{\begin{tabular}[c]{@{}c@{}}PTV \\ \textit{V}100\%\end{tabular}} & \textbf{\begin{tabular}[c]{@{}c@{}}PTV \\ \textit{V}150\%\end{tabular}} & \textbf{\begin{tabular}[c]{@{}c@{}}CTV \\ \textit{V}100\%\end{tabular}} & \textbf{\begin{tabular}[c]{@{}c@{}}CTV \\ \textit{V}150\%\end{tabular}} & \textbf{\begin{tabular}[c]{@{}c@{}}URE \\ \textit{V}150\%\end{tabular}} & \textbf{\begin{tabular}[c]{@{}c@{}}REC \\ \textit{V}50\%\end{tabular}} &
\textbf{\#N} & \textbf{\#S} \\ 
\hline
\textbf{$SA_{seattle}$} & $92.9 \pm 2.2$ & $47.6 \pm 5.1$ & $98.2 \pm 1.3$ & $55.2 \pm 7.2$ & $9.6 \pm 7.9$ & $14.8 \pm 4.6$ & $30 \pm 4$ & $100 \pm 13$\\\noalign{\smallskip}\hline

\textbf{$jSDL$} & $94.2 \pm 3.4$ & $53.6 \pm 5.6$ & $98.5 \pm 2.0$ & $61.9 \pm 8.3$ & $22.7 \pm 19.0$ & $16.8 \pm 7.7$ & $24 \pm 3$ & $112 \pm 16$\\\noalign{\smallskip}\hline

\textbf{$SA_{genN}$} & $94.8 \pm 1.7$ & $51.0 \pm 4.2$ & $98.9 \pm 1.1$ & $58.3 \pm 5.4$ & $6.2 \pm 5.6$ & $17.3 \pm 5.9$ & $28 \pm 3$ & $106 \pm 16$\\\noalign{\smallskip}\hline

\textbf{$TP\text{-}GAN$} & $\mathbf{94.6 \pm 3.9}$ & $\mathbf{55.0 \pm 11.9}$ & $\mathbf{97.8 \pm 2.5}$ & $\mathbf{60.8 \pm 13.7}$ & $\mathbf{7.8 \pm 11.9}$ & $\mathbf{17.6 \pm 5.4}$ & $\mathbf{24 \pm 3}$ & $\mathbf{109 \pm 17}$\\\noalign{\smallskip}\hline

\textbf{$TP\text{-}GAN+SA$} & $\mathbf{95.9 \pm 1.6}$ & $\mathbf{53.0 \pm 3.5}$ & $\mathbf{98.8 \pm 0.9}$ & $\mathbf{59.1 \pm 5.0}$ & $\mathbf{4.5 \pm 3.08}$ & $\mathbf{16.7 \pm 4.4}$ & $\mathbf{27 \pm 3}$ & $\mathbf{107 \pm 15}$\\\noalign{\smallskip}\hline

\textbf{$Actual$} & $96.9 \pm 1.2$ & $55.9 \pm 4.0$ & $99.4 \pm 0.7$ & $62.1 \pm 4.9$ & $3.3 \pm 5.2$ & $17.1 \pm 4.4$ & $24 \pm 3$ & $110 \pm 17$\\
\end{tabular}%
}
\end{table}

\begin{figure}[t]
\includegraphics[width=1.0\textwidth]{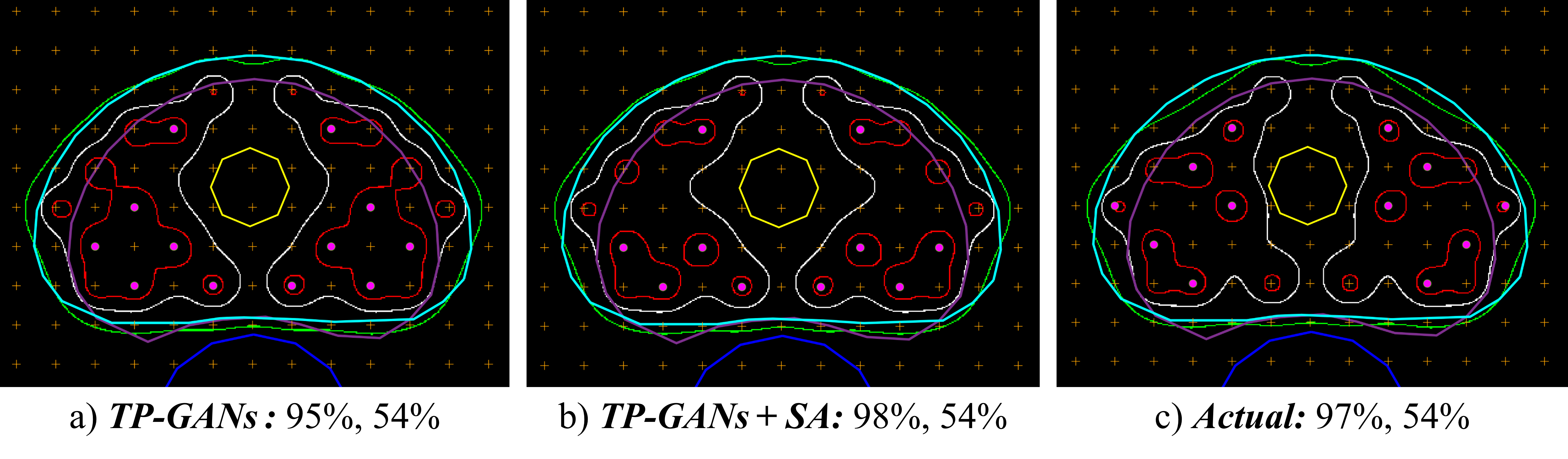}
\caption{Seed plans of a test patient generated using: a) $TP\text{-}GAN$, b) $TP\text{-}GAN+SA$, and c) Actual plan. For each technique, the numbers on the bottom of each figure are the respective PTV \textit{V}100\% and \textit{V}150\% of the total plans. Here, the template is represented by orange (+), seeds by pink dots, urethra by yellow margin, rectum by blue margin, CTV by purple margin, PTV by cyan margin, \textit{V}100\% by green margin, \textit{V}150\% by white margin, and \textit{V}200\% by red margin.
}
\label{fig:2}     
\end{figure}

A paired \textit{t}-test was used to find statistically significant differrence between the planning techniques. From Table \ref{tab:1}, $SA_{seattle}$ has inferior performance among all other techniques. We suspect that this is because its initialization is based on a template that may not always converge to the optimal solution. Compared to all other techniques, $jSDL$ has the highest OAR toxicity as its objective function does not consider OAR dosage. The SA we use explicitly includes OAR dosimetry in its cost function while $TP\text{-}GAN$ implicitly learns to avoid OAR from training data. Although, technique $SA_{genN}$ results are similar to those of $TP\text{-}GAN$, the latter can produce results instantly in less than 3 seconds compared to all other techniques from the table which can take from 2.5-30 mins/plan. Even though $TP\text{-}GAN$ can produce favourable plans on its own, its standard deviation across the metrics are high in general. Because of the limited dataset and possibility of multiple optimal solutions, this technique works best on similar prostate shapes it has seen in the training dataset. A general trend observed for $TP\text{-}GAN$ is that the number of seeds is overestimated for smaller prostate volumes and underestimated for larger volumes; resulting in over/under dose coverage on smaller/larger prostates. Adding SA tuning improves this significantly resulting in $TP\text{-}GAN+SA$ being the most comparable to manual plans (\textit{p}$<$0.05 for most metrics when compared with other techniques). This increases the planning duration from $<3$ seconds to an average of 2.5 mins/plan, which is still significantly lower than average manual planning times of 20 mins. Fig.~\ref{fig:2} shows results of generated plans using $TP\text{-}GAN$ and $TP\text{-}GAN+SA$ vs the actual prehistorical plan for a test patient. 
\begin{table}[t]
\centering
\caption{Mean and standard deviation of the metrics on different ablation settings.}
\label{tab:2}
\resizebox{0.9\textwidth}{!}{%
\begin{tabular}{ccc|c|c|c|c|}
\hline
\multicolumn{3}{|c||}{Ablation Settings} & \multirow{2}{*}{AUC} & \multirow{2}{*}{Dice coeff} & \multirow{2}{*}{Adj Seeds} & \multirow{2}{*}{Seed Diff} \\ \cline{1-3}
\multicolumn{1}{|c|}{Needle Plan} & \multicolumn{1}{c|}{Augmentation} & \multicolumn{1}{c||}{${L_{adj~seeds}}$} &  &  &  &  \\ \hline
\multicolumn{1}{|c|}{\ding{55}} & \multicolumn{1}{c|}{\ding{55}} & \multicolumn{1}{c||}{\ding{55}} & $89.1\% \pm 3.2\%$ & $34.6\% \pm 9.4\%$ & $18 \pm 7$ & $25 \pm  9$ \\
\multicolumn{1}{|c|}{\ding{51}} & \multicolumn{1}{c|}{\ding{55}} & \multicolumn{1}{c||}{\ding{55}} & $97.2\% \pm 2.2\%$ & $71.1\% \pm 9.1\%$ & $50 \pm 18$ & $23 \pm  9$ \\
\multicolumn{1}{|c|}{\ding{51}} & \multicolumn{1}{c|}{\ding{51}} & \multicolumn{1}{c||}{\ding{55}} & $97.4\% \pm 2.1\%$ & $73.4\% \pm 11.1\%$ & $30 \pm 15$ & $13 \pm  6$ \\
\multicolumn{1}{|c|}{\ding{51}} & \multicolumn{1}{c|}{\ding{51}} & \multicolumn{1}{c||}{\ding{51}} & $97.8\% \pm 2.0\%$ & $72.3\% \pm 14.1\%$ & $4 \pm 5$ & $6 \pm 4$ \\\hline
\end{tabular}%
}
\end{table}

\subsection{Ablation study}
We analyzed the importance of the different components of TP-GAN which includes the use of: needle plans in the input, augmentation, and $Adjacent~seed$ loss. All the different configurations, as indicated in Table \ref{tab:2}, were trained using the same hyper-parameters and was evaluated on the validation test. We compared the area under the receiver operating characteristic curve (AUC), Dice coefficient, number of adjacent seeds, and the difference between the number of seeds from predicted and real plans. From Table \ref{tab:2}, a clear improvement is seen with each additional feature. The adjacent seeds are low for no needle plan because the model is not predicting enough seeds as can be indicated by the very low Dice coefficient and seed difference. Adding needle plan improves the Dice coefficient significantly, but now with no constraint on the output, the model predicts a large number of adjacent seeds. Adding augmentation improves the results as the model overfits less and sees more variation in the input- indicated by the increase in the Dice coefficient and decrease in adjacent seeds \& seed difference. Finally, adding the proposed loss keeps the Dice coefficient in the same range while significantly lowering the adjacent seeds and the seed difference in the prediction.

\section{Discussion \& Conclusion}
We proposed a novel end-to-end method called TP-GAN for the automatic generation of LDR-PB treatment plans. To our best knowledge, this is the first method using fully automatic feature extraction to learn implicit clinical factors which are not possible to manually determine. A novel loss function is proposed to penalize unacceptable seed placements. A large pool of retrospective data from 961 cases was used for training, validating, and testing the model. Comprehensive evaluation was made between our proposed model with other automatic approaches for seed plan generation using pertinent clinical measures. From the results, we suggest the use $TP\text{-}GAN$ for rapid plan generation and the use of $TP\text{-}GAN + SA$ when further fine-tuning is required. Both of these methods can significantly save crucial time and resources for brachytherapy clinicians. Furthermore, such a method can be used for real-time plan generation for centres providing intra-operative LDR-PB treatment. 

\bibliographystyle{splncs04}
\bibliography{references}

\end{document}